\documentclass[letterpaper]{article} 
\usepackage{aaai2027}  
\usepackage[hyphens]{url}  
\usepackage{graphicx} 
\usepackage{natbib}  
\usepackage{caption} 
\usepackage{algorithm}
\usepackage{algorithmic}
\usepackage{amsmath}
\usepackage{newfloat}
\usepackage{listings}
\DeclareCaptionStyle{ruled}{labelfont=normalfont,labelsep=colon,strut=off} 
\floatstyle{ruled}
\newfloat{listing}{tb}{lst}{}
\floatname{listing}{Listing}

\usepackage{booktabs}

\title{Towards Bitstream-corrupted Harsh Visual Understanding: Through Bitstream Language Modeling as Robust Semantic Priors}

\author{
    Chaoran Huang,
    Fangcheng Li,
    Tianyi Liu,
    Wenyang Liu,
    Kejun Wu
}
\affiliations{}

\begin{document}

\maketitle

\begin{abstract}

Bitstream-corrupted Harsh Visual Understanding (BcHVU) aims to understand harshly degraded videos originally decoded from a severely corrupted bitstream in real-world multimedia communication.
The ill-posed nature of BcHVU poses a major challenge for existing vision models, as even subtle bitstream corruption can lead to irreversible pixel distortion and significant semantic loss.
To address these challenges in BcHVU, we propose Bitstream Language Modeling as Robust Semantic Priors (BLMSP), a framework for learning and injecting bitstream-native semantic cues.
Our proposed BLMSP framework learns to extract bitstream-native semantic cues by bitstream language modeling, and leverages them as priors by injecting into off-the-shelf vision models of BcHVU tasks.
Specifically, we present a Video Bitstream Byte Model (VBBM) that integrates byte-level modeling and cross-codec semantic distillation, enabling it to interpret robust semantics from byte sequences in multiple corrupted bitstream formats.
The learned bitstream semantics are leveraged as robust priors and fused into BcHVU model backbones for improving the quality of video restoration, captioning, and human pose estimation. 
To train BLMSP, we construct a large-scale multi-source Corrupted-bitstream Harsh-video Paired (CHP) dataset containing 607k corrupted bitstream segments and 287k paired harsh video clips.
Extensive experimental results show that the learned bitstream priors improve video restoration, captioning, and human pose estimation by 2.51 dB in PSNR, 0.20 in CIDEr, and 0.18 in PCK@0.2 on average, respectively.
These results demonstrate that corrupted bitstream can serve as robust semantic priors in solving pixel distortion and semantic loss in BcHVU. 

\end{abstract}


\section{Introduction}

Visual understanding supports applications such as intelligent surveillance, transportation, mobile devices, and online platforms~\cite{tang2025video}. An end-to-end video compression framework learns hierarchical temporal contexts within the coding pipeline~\cite{wu2025end}, but most downstream systems still decode compressed images or videos into pixels before applying vision models~\cite{ravanbakhsh2024deep}. Representative video models, including I3D, Video Swin Transformer, and VideoMAE, strengthen this decode-then-understand paradigm through spatiotemporal representation learning~\cite{carreira2017quovadis,liu2022videoswin,tong2022videomae}. However, the paradigm assumes that the compressed bitstream can be decoded correctly and that the reconstructed pixels retain sufficient visual and semantic information.

\begin{figure}[!t]
\centering
\includegraphics[width=\columnwidth]{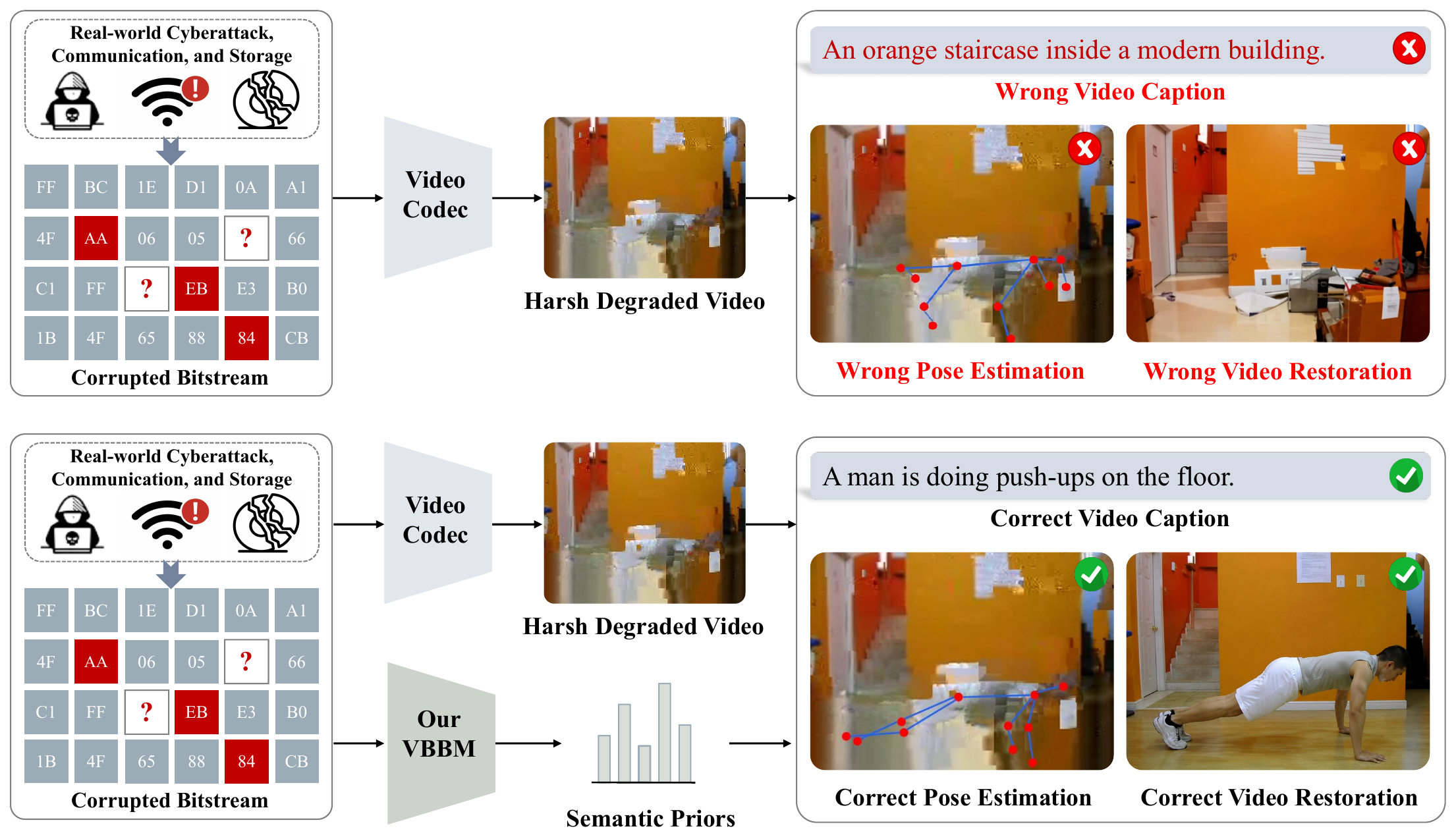}
\caption{Comparison between downstream processing without and with bitstream-native semantic priors.}
\label{fig:motivation}
\end{figure}

Compressed visual bitstreams can be corrupted by bit errors, packet loss, unstable communication links, or file truncation~\cite{liu2023bscvr,cheng2024grace,zou2026ntire}. Recent work therefore studies blind recovery from corrupted video bitstreams~\cite{liu2025towards}. Local errors may propagate during decoding and affect multiple spatial regions or consecutive frames. Mild corruption can still produce decoded content, but often with blocking artifacts, color distortion, missing regions, or damaged structures. Severe corruption can cause a standard decoder to reject the stream or return no valid visual output~\cite{cheng2024grace}. As illustrated by the top pathway in Fig.~\ref{fig:motivation}, these corruptions severely impair pixel-domain models, as the decoded pixels no longer provide reliable appearance information or semantic cues.

However, a corrupted bitstream is not necessarily semantically empty. File-fragment studies show that self-attention over bytes and neighboring sectors supports file-type recognition from raw fragments~\cite{wang2024intra}, while image-bitstream fragments can also be classified with language models~\cite{li2025ibfc}. Byte-level models can further learn directly from raw file bytes without modality-specific tokenization~\cite{horton2024bytes}. Recent approaches extend this idea to action recognition directly from corrupted JPEG bytes~\cite{li2026bitstreamactionrecognitionbyte}, semantic understanding of corrupted bitstreams with adaptive-modal language models~\cite{wu2026corrupted}, and pixel-free captioning from corrupted image bitstreams~\cite{liang2026cibic}. Compressed-domain methods exploit a related idea through codec-derived representations such as motion vectors and residuals~\cite{shou2019dmc,li2022end}, but they still require partial decoding or codec-specific syntax extraction~\cite{tian2023secure}. Their assumptions become unreliable when the bitstream is incomplete or severely corrupted.

We study Bitstream-corrupted Harsh Visual Understanding (BcHVU), the task of understanding videos whose compressed bitstreams are severely corrupted, resulting in heavily degraded or partially decodable visual inputs, and formulate their byte-level semantics as reusable priors for heterogeneous downstream models. To address this challenge, we learn semantic representations directly from compressed byte sequences and use them as robust priors for downstream visual understanding. As shown by the bottom pathway in Fig.~\ref{fig:motivation}, these priors provide complementary high-level semantics when standard decoding produces severely distorted content. When standard decoding fails, the downstream model still receives partial visual inputs recovered through permissive decoding, while the bitstream-native prior supplies missing semantic cues rather than replacing the visual branch. These observations motivate our framework, which treats corrupted bitstreams as an additional source of semantic evidence rather than merely as failed decoding inputs. 

Our main contributions are summarized as follows:

\begin{itemize}
\item We propose BLMSP, a corrupted-bitstream prior framework that learns bitstream-native action semantics from damaged byte sequences through VBBM, cross-codec semantic distillation, and task-specific prior injection, thereby providing high-level action cues when decoded pixels are unreliable.

\item We construct the Corrupted-bitstream Harsh-video Paired (CHP) dataset from three public benchmarks, covering MJPEG and H.264, nine corruption configurations, and protected and unprotected variants, thereby enabling controlled evaluation under diverse bitstream corruptions.

\item We evaluate BLMSP on video captioning, human pose estimation, and video restoration across codecs, datasets, and model families, demonstrating consistent improvements across all evaluation metrics.
\end{itemize}

\section{Methodology}

\begin{figure*}[!t]
\centering
\includegraphics[width=1.0\textwidth]{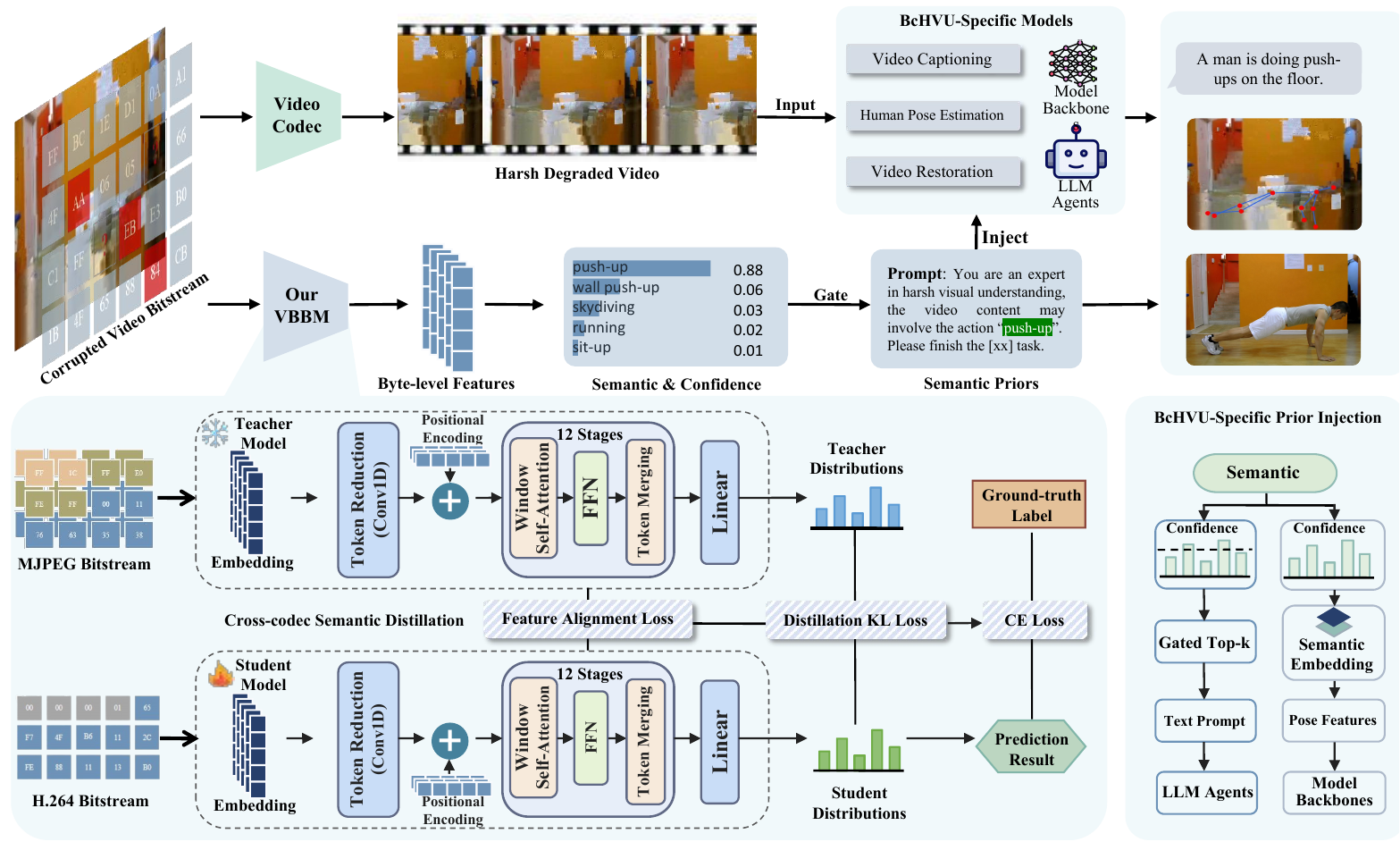}
\caption{
Overview of the proposed BLMSP framework for Bitstream-corrupted Harsh Visual Understanding (BcHVU). BLMSP learns semantic representations directly from corrupted video bitstreams and injects them as robust priors into downstream BcHVU-specific models. A cross-codec semantic distillation scheme further transfers knowledge from MJPEG to H.264 bitstreams through feature alignment and distribution-level supervision, enabling robust video captioning, human pose estimation, and video restoration under severe visual degradation.
}
\label{fig:framework}
\end{figure*}

\begin{figure}[!t]
\centering
\includegraphics[width=1.0\columnwidth]{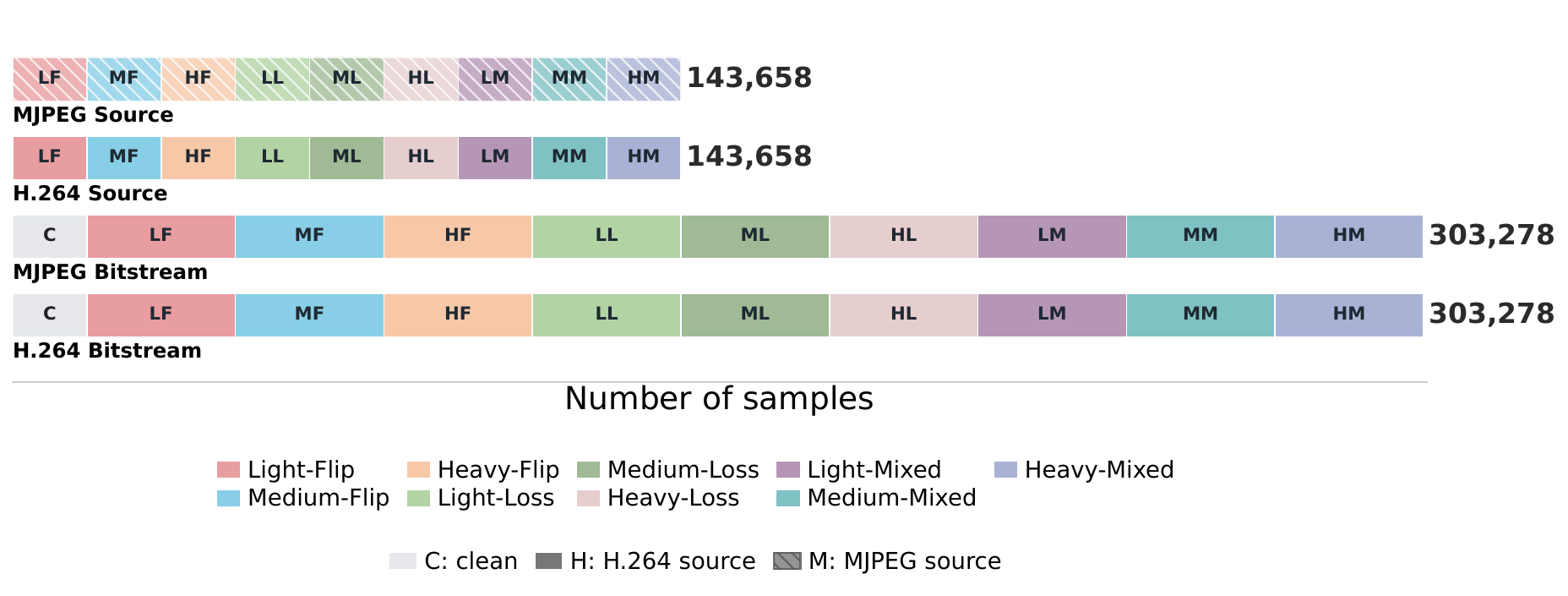}\par
{\footnotesize\textbf{(a)}}\par
\includegraphics[width=1.0\columnwidth]{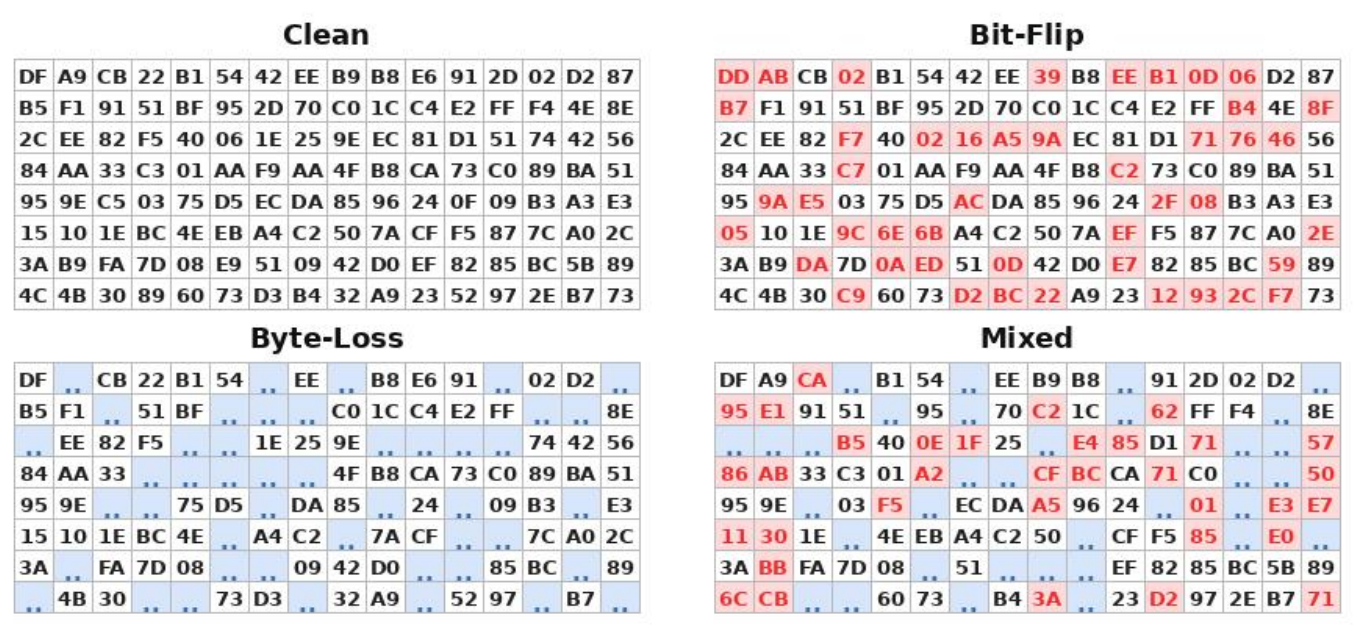}\par
{\footnotesize\textbf{(b)}}\par
\includegraphics[width=1.0\columnwidth]{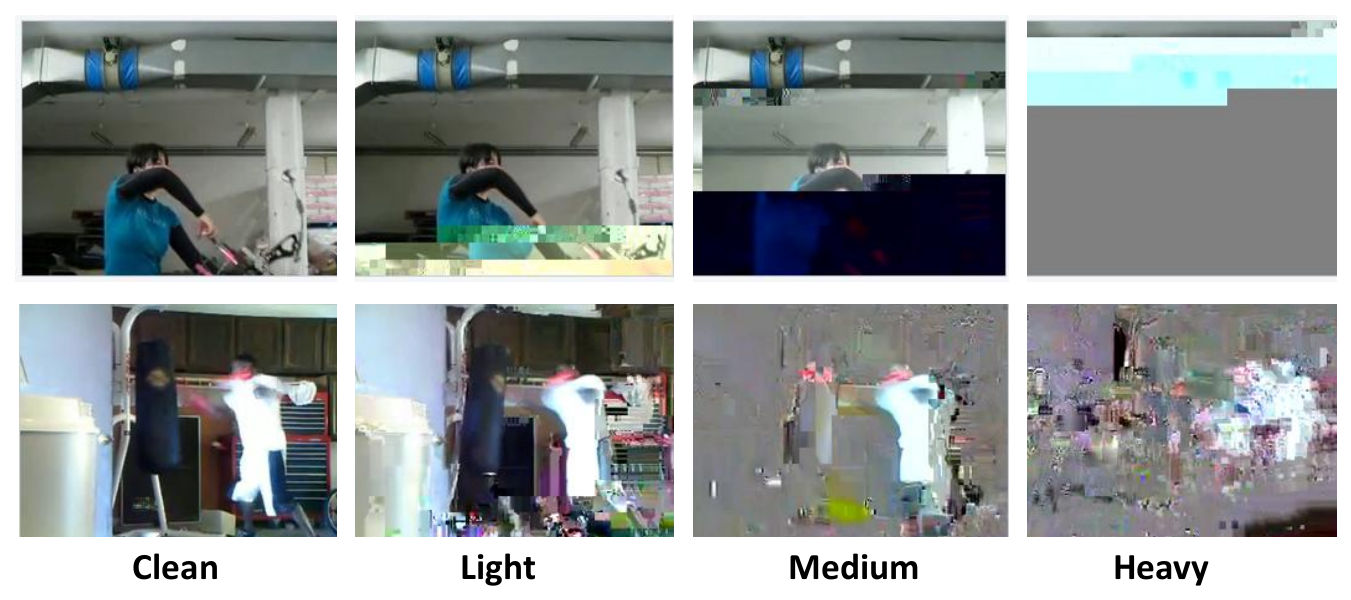}\par
{\footnotesize\textbf{(c)}}
\caption{Overview of the CHP dataset. (a) Sample distribution across corruption scenarios and source codecs. (b) Examples of clean, bit-flipped, byte-loss, and mixed corruption. (c) Decoded video frames with different corruption severities.}
\label{fig:corruption_dataset_decoding}
\end{figure}
\subsection{Overview}

As shown in Fig.~\ref{fig:framework}, BLMSP adopts a dual-branch framework for robust video understanding from corrupted bitstreams. The visual branch obtains partially recoverable frames through standard or permissive decoding and feeds them into task-specific models. In parallel, the Video Bitstream Byte Model (VBBM) directly processes corrupted byte sequences to extract semantic predictions without relying on reliable pixel reconstruction.

To improve semantic modeling for inter-frame codecs, VBBM is trained via cross-codec distillation, using an MJPEG teacher to supervise an H.264 student at both output and feature levels. At inference time, the predicted semantic distribution is converted into confidence-aware priors. These priors are injected as adaptive text prompts for video captioning and restoration, or as semantic embeddings for pose estimation.

\subsection{Bitstream Language Modeling}

Bitstream language modeling extracts semantic priors directly from corrupted compressed byte sequences without relying on reliable pixel reconstruction. We adopt ByteFormer~\cite{horton2024bytes} as the byte-level backbone to model coding patterns, byte dependencies, and codec syntax. For H.264, whose semantics are distributed across inter-frame prediction dependencies, we employ cross-codec knowledge distillation~\cite{hinton2015distilling} with an MJPEG teacher. MJPEG encodes frames independently, whereas H.264 uses inter-frame prediction~\cite{poynton2003jpeg,wiegand2003overview}. Let \(b^{m}\) and \(b^{h}\) denote the MJPEG and H.264 bitstreams generated from the same video clip with matched temporal sampling, spatial augmentation, and corruption settings. The frozen teacher \(F_{\mathrm{m}}\) receives \(b^{m}\), while the student \(F_{\mathrm{h}}\) receives \(b^{h}\).

The student is optimized with supervised classification, output-level distillation, and intermediate feature alignment:
\begin{equation}
\mathcal{L}_{\text{total}}
=
\mathcal{L}_{\text{CE}}
+
\mathcal{L}_{\text{KL}}
+
\mathcal{L}_{\text{feat}}.
\label{eq:total-loss}
\end{equation}
Here, \(\mathcal{L}_{\text{CE}}\) is the cross-entropy loss between the student prediction and the ground-truth action label, \(\mathcal{L}_{\text{KL}}\) is the temperature-scaled KL divergence between teacher and student output distributions, and \(\mathcal{L}_{\text{feat}}\) aligns the final intermediate representations. The combined objective encourages the H.264 model to preserve task-discriminative semantics while adapting to byte patterns induced by inter-frame compression.

\subsection{Bitstream-native Semantic Prior Injection}
  
Bitstream corruption can damage decoded frames through artifacts, missing regions, structural distortion, or undecodable segments. When standard decoding fails, permissive decoding tolerates damaged syntax and retains recoverable frames or regions as a partial visual input. VBBM then extracts semantic information directly from the corrupted stream and provides it to downstream models as a semantic prior. Given a corrupted bitstream \(b^{c}\), VBBM produces class logits
\begin{equation}
\mathbf{z}^{c}=F_{\mathrm{VBBM}}(b^{c}),
\label{eq:class-logits}
\end{equation}
which are converted into class probabilities with a temperature-scaled softmax:
\begin{equation}
\mathbf{p}^{c}_{i}
=
\frac{\exp\left(\mathbf{z}^{c}_{i}/\tau\right)}
{\sum_{j=1}^{C}\exp\left(\mathbf{z}^{c}_{j}/\tau\right)},
\label{eq:class-probabilities}
\end{equation}
where \(C\) denotes the number of semantic classes and \(\tau\) is the temperature parameter. Compared with a single Top-1 prediction, \(\mathbf{p}^{c}\) preserves inter-class similarity and prediction uncertainty. We further introduce a Top-\(k\) gating strategy, where the value of \(k\) is adaptively determined according to the Top-1 confidence score: a higher confidence selects fewer candidate classes, while a lower confidence retains more candidates to preserve semantic ambiguity.

The extracted semantic prior is injected into downstream tasks through task-specific pathways. For video captioning and video restoration, the Top-\(k\) semantic classes and their confidence scores are converted into confidence-aware text prompts and provided to LLMs to guide semantic description generation and visual reconstruction. For human pose estimation, the semantic distribution is transformed into semantic embeddings and fused with visual features within the pose estimation model to improve pose prediction under degraded inputs. In this way, the bitstream-derived prior complements the recovered visual information rather than replacing the visual branch.

\subsection{CHP Dataset Construction}

Despite recent progress in video-related datasets, publicly available benchmarks containing paired MJPEG and H.264 bitstreams of identical video content remain scarce, especially when controlled bitstream corruption is considered. To address this limitation, we construct CHP, a dedicated dataset that enables systematic evaluation across coding formats and corruption conditions.

We construct CHP from three public benchmarks: UCF101~\cite{soomro2012ucf101}, Penn Action~\cite{zhang2013actemes}, and Sub-JHMDB~\cite{jhuang2013towards}. UCF101 contains 13,320 clips from 101 action classes and is used for byte-level action classification, video captioning, and video restoration. Penn Action contains 2,326 videos from 15 action classes with annotations for 13 keypoints, and Sub-JHMDB contains 316 videos from 12 action classes with annotations for 15 keypoints; these two datasets support pose estimation. Figure~\ref{fig:corruption_dataset_decoding} visualizes the dataset composition, representative byte-level corruption patterns, and the resulting decoded frames across corruption severities. For UCF101, we use Qwen-VL-Max~\cite{alibaba2026qwenvlmax} to generate candidate descriptions from the original videos and retain three to five quality-filtered references for each validation sample for metric computation.

Corrupted MJPEG and H.264 streams are generated by adapting the Real-world Bitstream Corruption Simulator (RBCS) proposed in~\cite{li2026bitstreamactionrecognitionbyte}. In our adaptation, each bitstream is divided into fixed-length segments of \(S\) bytes, where each segment is selected with probability \(p\) and each byte within the selected segment is perturbed with probability \(q\). Smaller \(S\) and larger \(p\) or \(q\) indicate more severe corruption. The parameter \(\alpha\) controls the perturbation type, with \(\alpha=1\) for bit flipping, \(\alpha=0\) for byte deletion, and \(\alpha=0.5\) for an equal mixture. We define three corruption levels (light, medium, and heavy) for each perturbation type, resulting in nine configurations. To extend RBCS across codecs and our task setting, each configuration contains a protected variant preserving essential structural information and an unprotected variant without such protection. The unprotected variants are used for VBBM training to avoid reliance on residual codec-specific cues and encourage semantic learning directly from corrupted byte sequences. Table~\ref{tab:h264_perturbation} summarizes the parameter settings.

\begin{table}[t]
\centering
\begin{tabular*}{\linewidth}{@{\extracolsep{\fill}} l c c c c }
\toprule
\textbf{Scenario} & \textbf{$S$} & \textbf{$p$} & \textbf{$q$} & \textbf{$\alpha$} \\
\midrule
Light-Flip   & 512 & 0.05 & 0.15 & 1.0 \\
Medium-Flip  & 256 & 0.10 & 0.25 & 1.0 \\
Heavy-Flip   & 128 & 0.15 & 0.35 & 1.0 \\
\midrule
Light-Loss   & 512 & 0.05 & 0.15 & 0.0 \\
Medium-Loss  & 256 & 0.10 & 0.25 & 0.0 \\
Heavy-Loss   & 128 & 0.15 & 0.35 & 0.0 \\
\midrule
Light-Mixed  & 512 & 0.05 & 0.15 & 0.5 \\
Medium-Mixed & 256 & 0.10 & 0.25 & 0.5 \\
Heavy-Mixed  & 128 & 0.15 & 0.35 & 0.5 \\
\bottomrule
\end{tabular*}
\caption{Bitstream corruption parameters used to construct CHP. \(S\) is the segment length in bytes; \(p\) is the segment-selection probability; \(q\) is the per-byte perturbation probability; and \(\alpha\) controls the corruption type.}
\label{tab:h264_perturbation}
\end{table}

\begin{figure}[t]
    \centering
    \includegraphics[width=\linewidth]{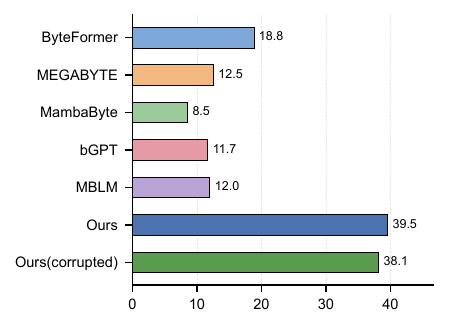}
    \caption{Top-1 action recognition accuracy of byte-level models on clean and corrupted UCF101 H.264 bitstreams. Existing methods are evaluated on clean bitstreams, while our model is evaluated under both clean and corrupted bitstream conditions.}
    \label{fig:byte_model_top1}
\end{figure}

\section{Experiments}
\subsection{Experimental Settings}

\begin{table*}[p]
\centering

\begingroup

\newcommand{\tabledelta}[1]{\hspace{0.08em}(#1)}
\newcommand{\withprior}[2]{\textbf{#1}\tabledelta{#2}}

\small
\renewcommand{\arraystretch}{0.97}

{
\setlength{\tabcolsep}{8pt}

\makebox[\textwidth][c]{%
\begin{tabular}{@{}lll*{6}{c}@{}}
\toprule
& & &
\multicolumn{3}{c}{PCK@0.2$\uparrow$} &
\multicolumn{3}{c}{PCK@0.1$\uparrow$} \\
\cmidrule(lr){4-6}
\cmidrule(lr){7-9}

Codec & Model & Dataset
& w/o prior & w/ prior & Gain (\%)
& w/o prior & w/ prior & Gain (\%) \\
\midrule

H.264 & HRNet-W48 & Penn Action
& 0.34 & \withprior{0.38}{+0.04} & +11.8\%
& 0.14 & \withprior{0.23}{+0.09} & +64.3\% \\

H.264 & RTMPose-M & Penn Action
& 0.38 & \withprior{0.43}{+0.05} & +13.2\%
& 0.17 & \withprior{0.20}{+0.03} & +17.6\% \\

H.264 & ViTPose-S & Penn Action
& 0.38 & \withprior{0.54}{+0.16} & +42.1\%
& 0.18 & \withprior{0.33}{+0.15} & +83.3\% \\

H.264 & LiteHRNet-18 & Penn Action
& 0.23 & \withprior{0.40}{+0.17} & +73.9\%
& 0.11 & \withprior{0.20}{+0.09} & +81.8\% \\

H.264 & MobileNetV2-HM & Penn Action
& 0.30 & \withprior{0.49}{+0.19} & +63.3\%
& 0.14 & \withprior{0.23}{+0.09} & +64.3\% \\

\midrule

H.264 & HRNet-W48 & Sub-JHMDB
& 0.60 & \withprior{0.82}{+0.22} & +36.7\%
& 0.29 & \withprior{0.58}{+0.29} & +100.0\% \\

H.264 & RTMPose-M & Sub-JHMDB
& 0.66 & \withprior{0.78}{+0.12} & +18.2\%
& 0.32 & \withprior{0.49}{+0.17} & +53.1\% \\

H.264 & ViTPose-S & Sub-JHMDB
& 0.69 & \withprior{0.83}{+0.14} & +20.3\%
& 0.37 & \withprior{0.57}{+0.20} & +54.1\% \\

H.264 & LiteHRNet-18 & Sub-JHMDB
& 0.26 & \withprior{0.48}{+0.22} & +84.6\%
& 0.12 & \withprior{0.21}{+0.09} & +75.0\% \\

H.264 & MobileNetV2-HM & Sub-JHMDB
& 0.47 & \withprior{0.75}{+0.28} & +59.6\%
& 0.21 & \withprior{0.38}{+0.17} & +81.0\% \\

\midrule

MJPEG & HRNet-W48 & Penn Action
& 0.22 & \withprior{0.29}{+0.07} & +31.8\%
& 0.06 & \withprior{0.12}{+0.06} & +100.0\% \\

MJPEG & RTMPose-M & Penn Action
& 0.31 & \withprior{0.34}{+0.03} & +9.7\%
& 0.12 & \withprior{0.15}{+0.03} & +25.0\% \\

MJPEG & ViTPose-S & Penn Action
& 0.28 & \withprior{0.47}{+0.19} & +67.9\%
& 0.09 & \withprior{0.17}{+0.08} & +88.9\% \\

MJPEG & LiteHRNet-18 & Penn Action
& 0.16 & \withprior{0.34}{+0.18} & +112.5\%
& 0.06 & \withprior{0.14}{+0.08} & +133.3\% \\

MJPEG & MobileNetV2-HM & Penn Action
& 0.18 & \withprior{0.29}{+0.11} & +61.1\%
& 0.07 & \withprior{0.18}{+0.11} & +157.1\% \\

\midrule

MJPEG & HRNet-W48 & Sub-JHMDB
& 0.43 & \withprior{0.80}{+0.37} & +86.0\%
& 0.16 & \withprior{0.42}{+0.26} & +162.5\% \\

MJPEG & RTMPose-M & Sub-JHMDB
& 0.59 & \withprior{0.79}{+0.20} & +33.9\%
& 0.19 & \withprior{0.43}{+0.24} & +126.3\% \\

MJPEG & ViTPose-S & Sub-JHMDB
& 0.56 & \withprior{0.88}{+0.32} & +57.1\%
& 0.25 & \withprior{0.52}{+0.27} & +108.0\% \\

MJPEG & LiteHRNet-18 & Sub-JHMDB
& 0.34 & \withprior{0.55}{+0.21} & +61.8\%
& 0.14 & \withprior{0.23}{+0.09} & +64.3\% \\

MJPEG & MobileNetV2-HM & Sub-JHMDB
& 0.41 & \withprior{0.76}{+0.35} & +85.4\%
& 0.15 & \withprior{0.36}{+0.21} & +140.0\% \\

\bottomrule
\end{tabular}%
}
}
\normalsize
\caption{Pose estimation performance without and with semantic prior
injection.}
\label{tab:pose-performance}
\medskip

\endgroup

\centering
\begingroup
\newcommand{\tabledelta}[1]{\hspace{0.08em}(#1)}
\newcommand{\withprior}[2]{\textbf{#1}\tabledelta{#2}}
\small
\renewcommand{\arraystretch}{0.97}

{
\setlength{\tabcolsep}{1.0pt}

\makebox[\textwidth][c]{%
\begin{tabular}{@{}ll*{10}{c}@{}}
\toprule
& &
\multicolumn{2}{c}{BLEU-1$\uparrow$} &
\multicolumn{2}{c}{BLEU-4$\uparrow$} &
\multicolumn{2}{c}{METEOR$\uparrow$} &
\multicolumn{2}{c}{ROUGE-L$\uparrow$} &
\multicolumn{2}{c}{CIDEr$\uparrow$} \\
\cmidrule(lr){3-4}
\cmidrule(lr){5-6}
\cmidrule(lr){7-8}
\cmidrule(lr){9-10}
\cmidrule(lr){11-12}

Codec & Model
& w/o prior & w/ prior
& w/o prior & w/ prior
& w/o prior & w/ prior
& w/o prior & w/ prior
& w/o prior & w/ prior \\
\midrule

H.264 & Qwen2.5-VL-3B
& 0.15 & \withprior{0.30}{+0.15}
& 0.07 & \withprior{0.13}{+0.06}
& 0.16 & \withprior{0.29}{+0.13}
& 0.12 & \withprior{0.25}{+0.13}
& 0.05 & \withprior{0.41}{+0.36} \\

H.264 & LLaVA-Video-7B
& 0.12 & \withprior{0.15}{+0.03}
& 0.07 & \withprior{0.09}{+0.02}
& 0.11 & \withprior{0.12}{+0.01}
& 0.07 & \withprior{0.11}{+0.04}
& 0.02 & \withprior{0.10}{+0.08} \\

H.264 & SmolVLM2-2.2B
& 0.16 & \withprior{0.19}{+0.03}
& 0.09 & \withprior{0.10}{+0.01}
& 0.12 & \withprior{0.17}{+0.05}
& 0.11 & \withprior{0.18}{+0.07}
& 0.03 & \withprior{0.19}{+0.16} \\

H.264 & Idefics2-8B
& 0.13 & \withprior{0.23}{+0.10}
& 0.06 & \withprior{0.11}{+0.05}
& 0.15 & \withprior{0.24}{+0.09}
& 0.11 & \withprior{0.20}{+0.09}
& 0.05 & \withprior{0.27}{+0.22} \\

H.264 & BLIP-Large-Grid
& 0.17 & \withprior{0.24}{+0.07}
& 0.10 & \withprior{0.12}{+0.02}
& 0.12 & \withprior{0.20}{+0.08}
& 0.12 & \withprior{0.19}{+0.07}
& 0.03 & \withprior{0.23}{+0.20} \\

\midrule

MJPEG & Qwen2.5-VL-3B
& 0.32 & \withprior{0.36}{+0.04}
& 0.15 & \withprior{0.16}{+0.01}
& 0.31 & \withprior{0.39}{+0.08}
& 0.27 & \withprior{0.32}{+0.05}
& 0.56 & \withprior{0.74}{+0.18} \\

MJPEG & LLaVA-Video-7B
& 0.20 & \withprior{0.27}{+0.07}
& 0.11 & \withprior{0.13}{+0.02}
& 0.21 & \withprior{0.28}{+0.07}
& 0.23 & \withprior{0.28}{+0.05}
& 0.34 & \withprior{0.49}{+0.15} \\

MJPEG & SmolVLM2-2.2B
& 0.20 & \withprior{0.29}{+0.09}
& 0.10 & \withprior{0.14}{+0.04}
& 0.19 & \withprior{0.29}{+0.10}
& 0.20 & \withprior{0.29}{+0.09}
& 0.21 & \withprior{0.54}{+0.33} \\

MJPEG & Idefics2-8B
& 0.22 & \withprior{0.27}{+0.05}
& 0.10 & \withprior{0.12}{+0.02}
& 0.22 & \withprior{0.27}{+0.05}
& 0.18 & \withprior{0.25}{+0.07}
& 0.23 & \withprior{0.37}{+0.14} \\

MJPEG & BLIP-Large-Grid
& 0.18 & \withprior{0.26}{+0.08}
& 0.10 & \withprior{0.13}{+0.03}
& 0.13 & \withprior{0.22}{+0.09}
& 0.12 & \withprior{0.21}{+0.09}
& 0.08 & \withprior{0.29}{+0.21} \\

\bottomrule
\end{tabular}%
}
}
\normalsize
\caption{Video captioning performance without and with semantic prior
injection.}
\label{tab:caption-performance}
\medskip

\endgroup

\centering
\begingroup
\newcommand{\tabledelta}[1]{\hspace{0.08em}(#1)}
\newcommand{\withprior}[2]{\textbf{#1}\tabledelta{#2}}
\small
\renewcommand{\arraystretch}{0.97}

{
\setlength{\tabcolsep}{4pt}

\makebox[\textwidth][c]{%
\begin{tabular}{@{}ll*{8}{c}@{}}
\toprule
& &
\multicolumn{2}{c}{PSNR $\uparrow$} &
\multicolumn{2}{c}{SSIM $\uparrow$} &
\multicolumn{2}{c}{LPIPS $\downarrow$} &
\multicolumn{2}{c}{DISTS $\downarrow$} \\
\cmidrule(lr){3-4}
\cmidrule(lr){5-6}
\cmidrule(lr){7-8}
\cmidrule(lr){9-10}

Codec & Model
& w/o prior & w/ prior
& w/o prior & w/ prior
& w/o prior & w/ prior
& w/o prior & w/ prior \\
\midrule

H.264 & Gemini 2.5 Flash Image
& 9.47 & \withprior{11.62}{+2.15}
& 0.18 & \withprior{0.24}{+0.06}
& 0.74 & \withprior{0.72}{-0.02}
& 0.40 & \withprior{0.37}{-0.03} \\

H.264 & Qwen-Image-Edit-Plus
& 9.27 & \withprior{11.66}{+2.39}
& 0.19 & \withprior{0.29}{+0.10}
& 0.78 & \withprior{0.73}{-0.05}
& 0.40 & \withprior{0.38}{-0.02} \\

H.264 & FLUX.1-Fill-dev
& 10.67 & \withprior{11.96}{+1.29}
& 0.08 & \withprior{0.10}{+0.02}
& 1.07 & \withprior{0.94}{-0.13}
& 0.59 & \withprior{0.50}{-0.09} \\

H.264 & Doubao-Seedream-4.5
& 10.08 & \withprior{11.55}{+1.47}
& 0.24 & \withprior{0.31}{+0.07}
& 0.65 & \withprior{0.58}{-0.07}
& 0.35 & \withprior{0.32}{-0.03} \\

H.264 & GPT Image 2
& 10.70 & \withprior{12.40}{+1.70}
& 0.34 & \withprior{0.36}{+0.02}
& 0.61 & \withprior{0.56}{-0.05}
& 0.32 & \withprior{0.28}{-0.04} \\

\midrule

MJPEG & Gemini 2.5 Flash Image
& 11.79 & \withprior{16.83}{+5.04}
& 0.40 & \withprior{0.59}{+0.19}
& 0.50 & \withprior{0.36}{-0.14}
& 0.30 & \withprior{0.23}{-0.07} \\

MJPEG & Qwen-Image-Edit-Plus
& 7.87 & \withprior{9.49}{+1.62}
& 0.23 & \withprior{0.27}{+0.04}
& 0.77 & \withprior{0.69}{-0.08}
& 0.42 & \withprior{0.37}{-0.05} \\

MJPEG & FLUX.1-Fill-dev
& 10.98 & \withprior{12.52}{+1.54}
& 0.06 & \withprior{0.08}{+0.02}
& 1.15 & \withprior{0.99}{-0.16}
& 0.64 & \withprior{0.56}{-0.08} \\

MJPEG & Doubao-Seedream-4.5
& 12.82 & \withprior{16.33}{+3.51}
& 0.51 & \withprior{0.62}{+0.11}
& 0.47 & \withprior{0.33}{-0.14}
& 0.27 & \withprior{0.22}{-0.05} \\

MJPEG & GPT Image 2
& 10.51 & \withprior{14.89}{+4.38}
& 0.38 & \withprior{0.59}{+0.21}
& 0.47 & \withprior{0.28}{-0.19}
& 0.29 & \withprior{0.22}{-0.07} \\

\bottomrule
\end{tabular}%
}
}
\normalsize
\caption{Video restoration performance without and with semantic prior
injection.}
\label{tab:prior_comparison}

\endgroup
\end{table*}

\begin{figure*}[!htbp]
\centering
\includegraphics[width=0.95\textwidth]{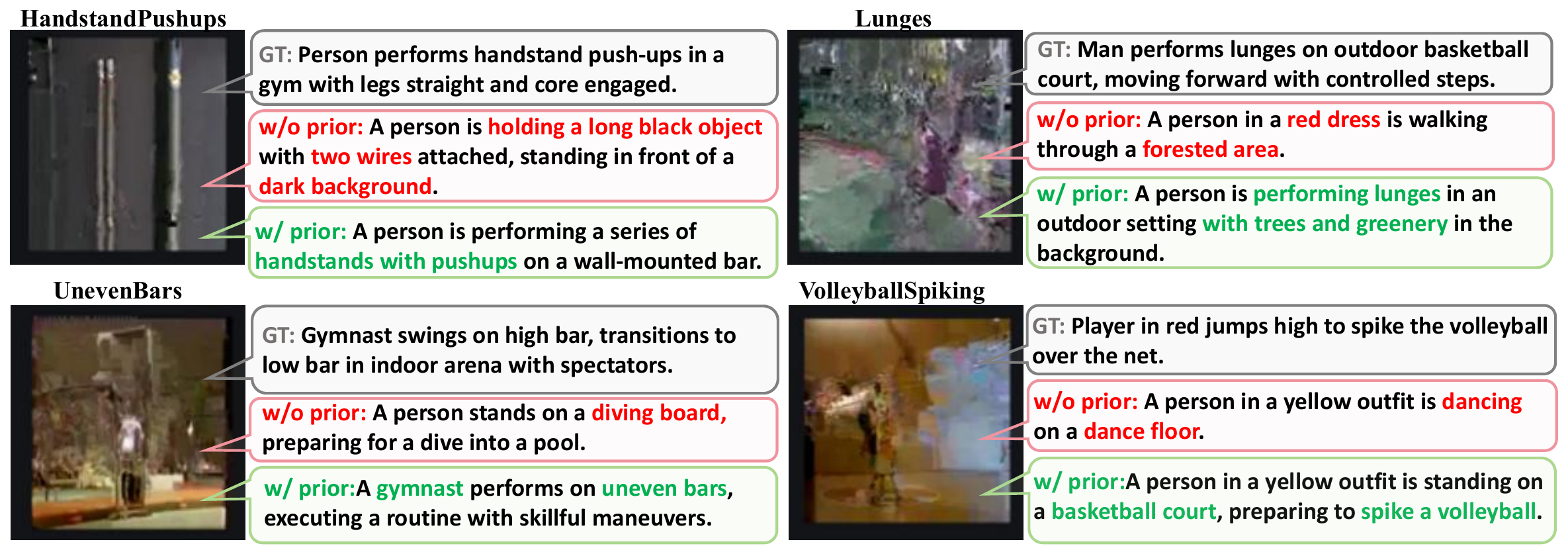}\par
{\footnotesize\textbf{(a)}}\par
\includegraphics[width=0.95\textwidth]{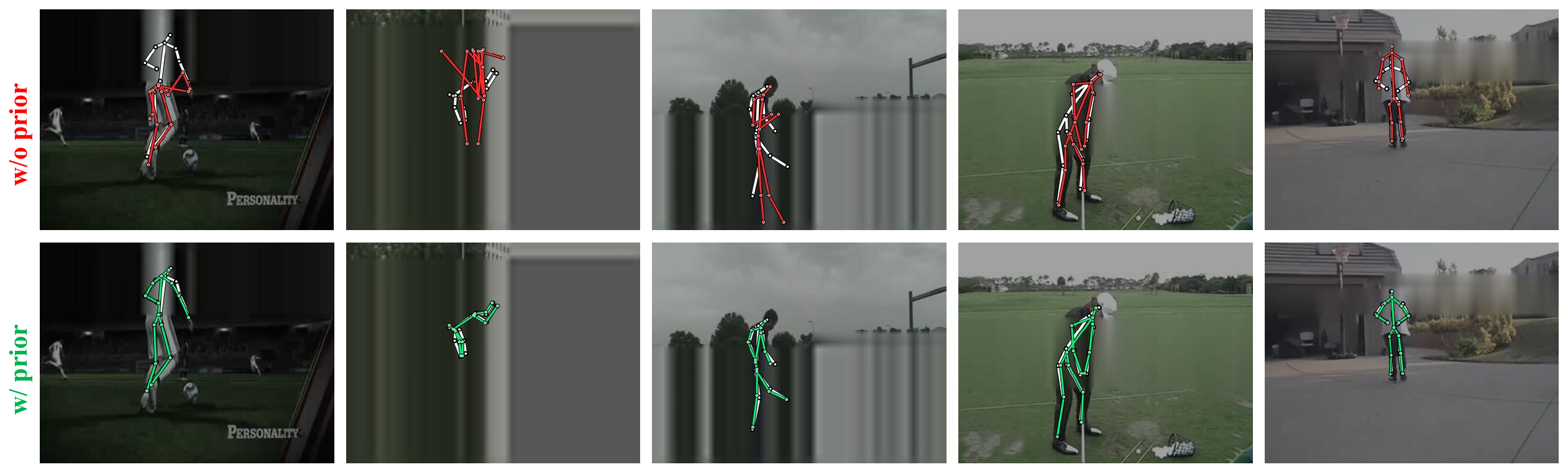}\par
{\footnotesize\textbf{(b)}}\par
\includegraphics[width=0.95\textwidth]{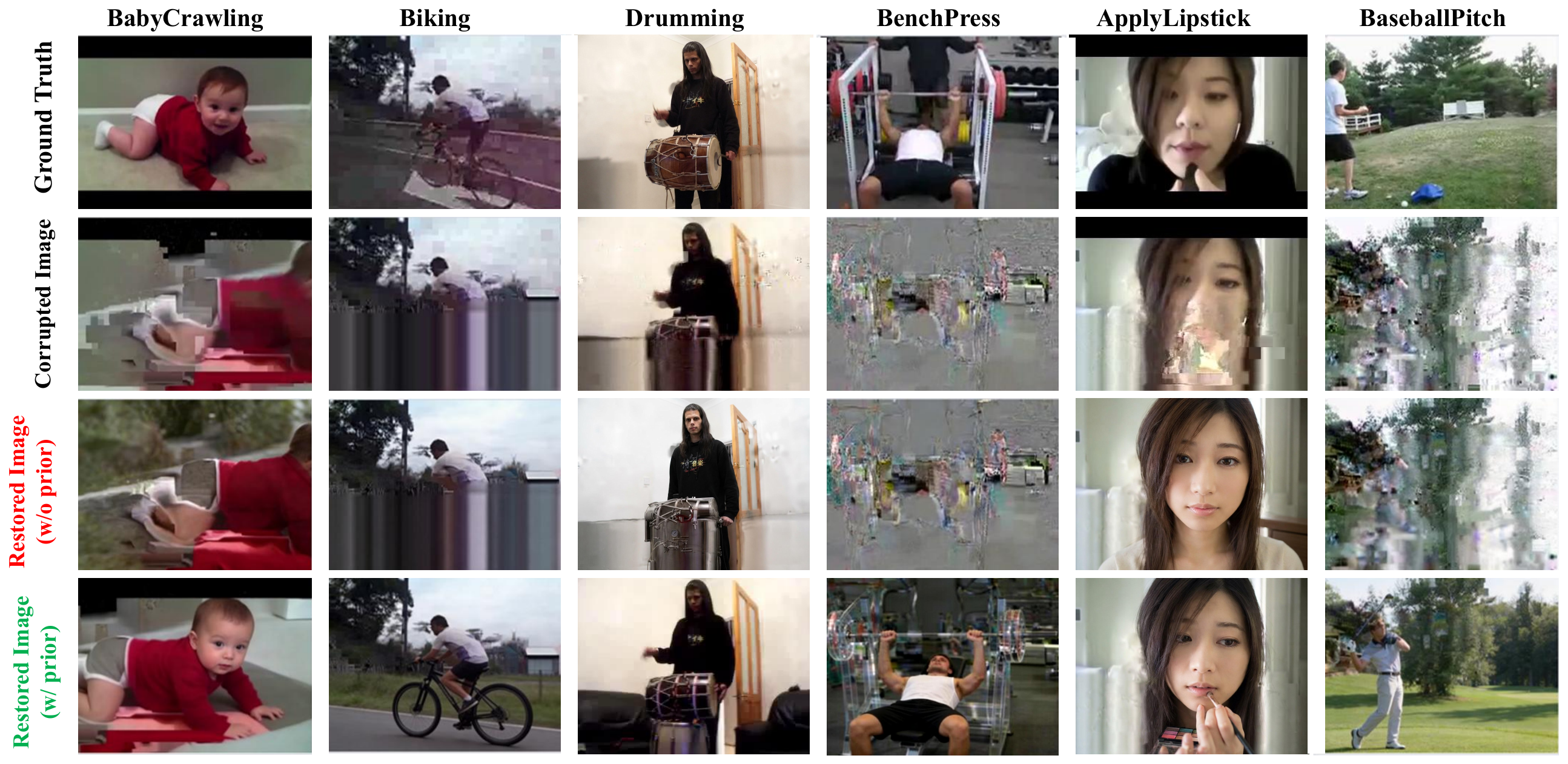}\par
{\footnotesize\textbf{(c)}}
\caption{Qualitative examples under bitstream corruption. (a) Video captioning compares ground-truth descriptions, captions without priors, and captions with priors. (b) Human pose estimation uses white, red, and green skeletons to denote ground truth, without priors, and with priors, respectively. (c) Video restoration compares ground-truth images, corrupted inputs, restorations without priors, and restorations with priors.}
\label{fig:qualitative_display}
\end{figure*}


We use a two-stage training strategy. First, we train an MJPEG bitstream classifier on UCF101 using four frames as input. The model is initialized with ImageNet JPEG-pretrained ByteFormer weights~\cite{horton2024bytes} and trained for 50 epochs. Next, we distill the MJPEG teacher model into an H.264 student model using 3-second video clips compressed at CRF values ranging from 25 to 30, together with pixel-level augmentation. The student model is trained for 200 epochs. After cross-codec distillation, the MJPEG teacher and H.264 student models are separately fine-tuned on the CHP dataset to obtain their respective Video Bitstream Byte Models (VBBMs). All models are trained on four NVIDIA RTX 4090 GPUs. The ByteFormer backbone is configured with a convolution kernel size of 8 and a window size of 128. For downstream tasks, the with-prior and no-prior models use identical corrupted inputs, decoding procedures, data splits, and training protocols. They differ only in whether the semantic prior produced by the corresponding VBBM is injected.

\subsection{Learning Bitstream-Native Robust Semantic Priors}

We first assess whether byte-level front ends can extract the action-semantic representations required by downstream models. For H.264 bitstreams, we compare direct supervised training, cross-codec distillation, and different byte-level models. Figure~\ref{fig:byte_model_top1} reports Top-1 recognition accuracy on clean UCF101 H.264 bitstreams for all compared methods, providing a controlled evaluation of semantic prior quality.

As shown in Figure~\ref{fig:byte_model_top1}, the proposed bitstream model achieves the highest Top-1 accuracy among all compared byte-level models on clean H.264 bitstreams. It improves over ByteFormer~\cite{horton2024bytes}, MEGABYTE~\cite{yu2023megabyte}, and MBLM~\cite{egli2025multiscale}, supporting the value of codec-aware representation learning over generic byte sequence modeling. Notably, even under corrupted H.264 bitstreams, the proposed model achieves higher Top-1 accuracy than the other compared byte-level models evaluated on clean bitstreams.

\subsection{Semantic-Prior-Injected Video Captioning}
We evaluate semantic-prior-injected video captioning on corrupted MJPEG and H.264 videos from UCF101. Table~\ref{tab:caption-performance} includes Qwen2.5-VL-3B~\cite{bai2025qwen25vl}, LLaVA-Video-7B~\cite{zhang2024llavavideo}, SmolVLM2-2.2B~\cite{marafioti2025smolvlm}, Idefics2-8B~\cite{laurencon2024idefics2}, and BLIP-Large-Grid~\cite{li2022blip}. For each model, we compare captions generated from decoded corrupted frames alone with captions generated using the action classes and confidence scores predicted by VBBM. Each sample is evaluated against three to five quality-filtered reference descriptions using BLEU-1/4~\cite{papineni2002bleu}, METEOR~\cite{banerjee2005meteor}, ROUGE-L~\cite{lin2004rouge}, and CIDEr~\cite{vedantam2015cider}. The reference descriptions are fixed for both comparison conditions, while the same base prompt template and decoding settings are used.

Table~\ref{tab:caption-performance} and Figure~\ref{fig:qualitative_display}(a) show improvements for every reported model--codec pair after semantic-prior injection. The gains across lexical-overlap metrics and CIDEr are consistent with the interpretation that action-level bitstream semantics help captioning models recover event information from degraded visual evidence.

\subsection{Semantic-Prior-Injected Human Pose Estimation}

We evaluate semantic-prior-injected human pose estimation on Penn Action and Sub-JHMDB using MJPEG and H.264 bitstreams. For each codec setting, VBBM first predicts an action prior and injects it into five pose estimators: HRNet-W48~\cite{sun2019deep}, RTMPose-M~\cite{jiang2023rtmpose}, ViTPose-S~\cite{xu2022vitpose}, LiteHRNet-18~\cite{yu2021litehrnet}, and MobileNetV2-HM~\cite{sandler2018mobilenetv2}. For RTMPose-M, the projected action prior is added to the keypoint-level hidden representation after the MLP projection and before the GAU module. For ViTPose-S, HRNet-W48, LiteHRNet-18, and MobileNetV2-HM, which use heatmap-based pose heads in our implementation, the projected prior is fused with the spatial pose features before the final heatmap prediction layer. Specifically, the action prior is mapped to the feature-channel dimension, broadcast over spatial dimensions, and added to the pose feature map before the final heatmap convolution.

We compare all estimators with and without the semantic prior under identical corrupted inputs and report PCK@0.2 and PCK@0.1. Table~\ref{tab:pose-performance} and Fig.~\ref{fig:qualitative_display}(b) show that across all evaluated pose models and codec settings, injecting the semantic prior consistently improves both metrics, indicating that action-level semantic information provides complementary contextual cues that help recover more accurate keypoints from degraded visual inputs.

\subsection{Semantic-Prior-Injected Video Restoration}

We evaluate semantic-prior-injected video restoration by corrupting MJPEG and H.264 bitstreams and decoding them into degraded visual inputs. The evaluated generative restoration models are Gemini 2.5 Flash Image~\cite{google2025gemini25flashimage}, Qwen-Image-Edit-Plus~\cite{alibaba2026qwenimageedit}, FLUX.1-Fill-dev~\cite{blackforestlabs2024flux}, Doubao-Seedream-4.5~\cite{bytedance2025seedream45}, and GPT Image 2~\cite{openai2026gptimage2}. Each model processes every degraded frame in two settings: restoration from the corrupted visual input alone and restoration with the action classes and confidence scores predicted by VBBM. PSNR and SSIM~\cite{wang2004ssim} measure pixel fidelity and structural similarity, while LPIPS~\cite{zhang2018unreasonable} and DISTS~\cite{ding2020image} measure perceptual and structure--texture quality. Table~\ref{tab:prior_comparison} reports the results for all five listed restoration models, with metric changes shown in parentheses. This experiment evaluates frame-level video restoration from degraded decoded inputs.

Table~\ref{tab:prior_comparison} and Fig.~\ref{fig:qualitative_display}(c) show that every reported model--codec pair improves in PSNR and SSIM and decreases in LPIPS and DISTS after prior injection, indicating that bitstream-native priors help recover more faithful and perceptually consistent visual content from corrupted inputs.

\section{Conclusion}

In this paper, we introduced BLMSP for Bitstream-corrupted Harsh Visual Understanding (BcHVU). Unlike conventional decode-then-understand pipelines that rely solely on unreliable reconstructed pixels, BLMSP extracts semantic priors directly from corrupted bytes through VBBM and injects them into captioning, pose estimation, and video restoration models. By providing complementary action-level cues, BLMSP improves downstream performance under MJPEG and H.264 corruption. Its action-centric prior may not capture fine-grained or scene-level semantics, and its effectiveness under unseen codecs and natural transmission errors remains unexplored. Future work will investigate richer codec-agnostic priors and broader temporal understanding.

\bibliography{aaai2027}

\end{document}